\documentclass[twoside,11pt]{article}

\usepackage[abbrvbib, preprint]{jmlr2e}

\usepackage[utf8]{inputenc} % allow utf-8 input
\usepackage[T1]{fontenc}    % use 8-bit T1 fonts
\usepackage{hyperref}       % hyperlinks
\usepackage{url}            % simple URL typesetting
\usepackage{booktabs}       % professional-quality tables
\usepackage{amsfonts}       % blackboard math symbols
\usepackage{nicefrac}       % compact symbols for 1/2, etc.
\usepackage{microtype}      % microtypography
\usepackage{lipsum}
\usepackage{caption}
\usepackage{array}
\usepackage{gensymb}
\usepackage{adjustbox}
\usepackage[table]{xcolor}
\usepackage{amsmath}
\usepackage{algorithm}
\usepackage[noend]{algpseudocode}
\usepackage{adjustbox}
\usepackage{multirow}
\usepackage{colortbl}

\jmlrheading{1}{2020}{1-48}{4/00}{10/00}{mcshap}{Indranil Basu and Subhadip Maji}

\ShortHeadings{Multicollinearity Correction and Combined Feature Effect in Shapley Values}{Basu and Maji}
\firstpageno{1}

\begin{document}

\title{Multicollinearity Correction and Combined Feature Effect in Shapley Values}

\author{\name Indranil Basu \email basu.indranil@optum.com \\
       \addr Manager Data Science\\
       Optum UnitedHealth Group\\
       Bangalore, Karnataka, 560103
       \AND
       \name Subhadip Maji \email maji.subhadip@optum.com \\
       \addr Data Scientist\\
       Optum UnitedHealth Group\\
       Bangalore, Karnataka, 560103}

\editor{}

\maketitle

\begin{abstract}%   <- trailing '%' for backward compatibility of .sty file
Model interpretability is one of the most intriguing problems in most of the Machine Learning models, particularly for those that are mathematically sophisticated. Computing Shapley Values are arguably the best approach so far to find the importance of each feature in a model, at the row level. In other words, Shapley values represent the importance of a feature for a particular row, especially for Classification or Regression problems. One of the biggest limitations of Shapley vales is that, Shapley value calculations assume all the features are uncorrelated (independent of each other), this assumption is often incorrect. To address this problem, we present a unified framework to calculate Shapley values with correlated features. To be more specific, we do an adjustment (Matrix formulation) of the features while calculating Independent Shapley values for the rows. Moreover, we have given a Mathematical proof against the said adjustments. With these adjustments, Shapley values (Importance) for the features become independent of the correlations existing between them. We have also enhanced this adjustment concept for more than features. As the Shapley values are additive, to calculate combined effect of two features, we just have to add their individual Shapley values. This is again not right if one or more of the features (used in the combination) are correlated with the other features (not in the combination). We have addressed this problem also by extending the correlation adjustment for one feature to multiple features in the said combination for which Shapley values are determined. Our implementation of this method proves that our method is computationally efficient also, compared to original Shapley method.
\end{abstract}


\section{Introduction}
The ability to correctly interpret a prediction model’s output is extremely important. It engenders appropriate user trust, provides insight into how a model may be improved, and supports understanding of the process being modeled. Shapley values serve this purpose to a great extent with the biggest limitation that in case the used features are correlated, then Shapley calculations do not consider that. For example, for a given data point, if two features $X_1$ and $X_2$ with no correlation are having Shapley values 0.2 and 0.3, then with significant correlation, these Shapley values will definitely change. Simple reason is that, when we calculate the importance of $X_1$ (According to Shapley method), we replace $X_1$ (in a modified Shapley technique, we remove $X_1$) to find the effect on the other features including $X_2$. Now, if $X_1$ and $X_2$ are having high correlation, then replacement or removal of $X_1$ will increase the importance of $X_2$. Such effect is undesirable in calculation the right importance of $X_2$ (and vice versa, while calculating the importance of $X_1$). Therefore, we adjust $X_2$ with a linear modification (say $X_2^\prime$) so that $X_2^\prime$ would not have any correlation with $X_1$, thereby nullifying the correlation effect of $X_1$ while calculating the importance of $X_1$ by removing/replacing $X_1$. This method would be extended while calculating the combined importance of $X_1, X_2$ (or any other combination of any size).

Again, if we are supposed to find the combined importance of ($X_1, X_2$) for the proposed model, Combined Shapley value of ($X_1, X_2$) cannot be a simple summation of Shapley ($X_1$) and Shapley($X_2$) when $X_1$ and $X_2$ are individually correlated with other features, say $X_3, X_4, …., X_k$ (Let’s say there are K features in the model). In other words, combined effect of ($X_1, X_2$) should be less than Shapley ($X_1$) + Shapley ($X_2$). In the example above, this combined effect $< (0.2 + 0.3) = 0.5$, as proposed by the original Shapley method. To address this issue and quantify the combined effect, first we find the independent Shapley values (Importance values) of the features using the technique mentioned above. Then, we do the modification for the combined effect as an enhancement of the same method. Suppose, we want to modify $X_3$ to $X_3^\prime$ where $X_3^\prime$ would be uncorrelated with both $X_1$ and $X_2$ (features that are part of the said combination). So, we have added a linear adjustment factor with $X_3$ (to make $X_3^\prime$) and this factor is a function of covariance of $X_3$ with $X_1$ and $X_2$ individually.

In general, if there are $k$ features, we present a linear adjustment based on a combination of $p$ features on rest of the features (i.e. rest $k – p$ features) in a matrix formulation, given below. It is mentionable that while calculating the feature combination effect of m features, we have to find independent feature importance values using the method and formulation stated in the first step.

Moreover, we present a detailed Mathematical proofs regarding how we have reached the matrix formulations while adjusting for the effect of combination $X_1, X_2, …., X_k$ upon rest of the features. It is quite intuitive that the matrix formulation representing the linear adjustment is pretty much similar for rest of the features. A very practical assumption for this Mathematical derivation is that, all the correlations are linear in nature. For most of the real time scenario where features are extracted properly, this assumption should hold good. Necessary Mathematical derivation considering those non linear correlation effects are also possible without much hassle.

Biggest advantage of this modification addressing the correlation effects is that, linear correlations among the features in a dataset are very much common for the datasets evolving from most of the domains, e.g. Healthcare, Retail, Telecom, E-commerce. For example, importance values of smoking and drinking are $0.4$ and $0.3$ calculated from original Shapley method respectively for a possible early heart attack. It is quite obvious that people who smoke and the people who drink are very much correlated. Then, Shapley cannot find their independent importance values correctly in respect of possible heart attach due to smoking and drinking. Our technique definitely helps an Analyst or Healthcare professional under the independent effects of smoking and drinking. Also, the combined effect of smoking and drinking cannot be the sum of $0.4$ and $0.3$ as depicted in the original Shapley method. Our technique rightly addresses the same.

\section{Related Work}
Correlation effects can be handled only with a few methods mentioned in chapter 5 of the book, A Guide for Making Black Box Models Explainable by \cite{molnar2019}. First one is that, permute correlated features together and get one mutual Shapley value for them. Second one is to determine conditional sampling: Features are sampled conditional on the features that are already in the team. In both the approaches, we cannot find feature importance values independent of each other. 

\section{Our Approach}
For a given data (\textbf{X}, $y$), if two features $X_j$ and $X_k$ (where $j \neq k$) are linearly correlated, then we propose three novel ideas to correct the shapley output of a feature based on the feature types of the data available. $X_j$ is the $j$-th feature where $j = 1, 2, 3, ..., m$ and $m$ is the total number of features. $y$ is the response variable.

The main idea of multi-collinearity correction while calculating shapely values of feature $X_j$ for datapoint $i$, is that we remove the correlation effect of $X_j$ from all of the other features $X_k$, where $k = 1, 2, 3, ..., m$, and $j \neq k$. We have tested this algorithm for  shapely value calculation including multiple machine learning models. 

\subsection{Numeric Predictors Only}
Unlike shapley values, as the Multi-collinearity Corrected (MCC) Shapley values are not additive, we have to calculate MCC shapley value for a specific feature combinations to get the same. We break this subsection into two parts, where in the first part we discuss about the calculation of MCC shapley values for individual features and in the next part we discuss the same for the combination of two or more features. 

\subsubsection{MCC Shapley Values for Individual Features}
Assume for a dataset the correlation of $X_j$ with other features $X_1, X_2, ...,X_{j-1}, X_{j+1}, ..., X_m$ are $c_{j1}, c_{j2}, ..., c_{j(j-1)}, c_{j(j+1)}, ..., c_{jm}$ respectively. If we are interested in calculating the shapely value of $X_j$, we add one Adjustment Factor ($AF_k$) with $X_k$, where $k = 1, 2, ..., j-1, j+1, ..., m$, while we randomize (or remove) $X_j$ in the shapley value calculation process so that,

\begin{equation}
	cor(X_j, X_k +  AF_k) = 0
	\label{eq:indv_1}
\end{equation}

Putting $AF_k = aX_j$ in the above equation and solving we get,

\begin{equation}
	AF_k = -\frac{cov(X_j, X_k)}{var(X_j)}X_j
	\label{eq:indv_2}
\end{equation}
The detailed steps from Equation \ref{eq:indv_1} to Equation \ref{eq:indv_2} have been shown in Appendix \ref{app:feat_1}. The reason of taking $AF_k$ as only a function of $X_j$, because we want to remove the correlation effect of $X_k$ only from $X_j$.

\subsubsection{MCC Shapley Values for Combination of Two or More Features}
\label{sec:mcc2}
Assume for a dataset the correlation of $X_i$ and $X_j$ with other features $X_k$, where $k = 1, 2, ..., m$ and $k \notin \{i, j\}$, are $c_{ik}$ and $c_{jk}$ respectively. If we are interested in calculating the shapely value of the combination of $X_i$ and $X_j$, we add one Adjustment Factor ($AF_k$) with $X_k$, while we randomize (or remove) $X_i$ and $X_j$ all together in the shapley value calculation process so that,

\begin{equation} 
	\label{eq:two_feat_1}
	\begin{split}
		cor(X_i, X_k +  AF_k) = 0 \\
		cor(X_j, X_k +  AF_k) = 0
	\end{split}
\end{equation}

Putting $AF_k = aX_i+bX_j$ in the above equation and solving we get,

\begin{equation} 
\label{eq:two_feat_2}
\begin{split}
a =  \frac{cov(X_i, X_k)var(X_j) - cov(X_j, X_k)cov(X_i, X_j)}{var(X_i)var(X_j) - (cov(X_i, X_j))^2}\\
b = \frac{cov(X_j, X_k)var(X_i) - cov(X_i, X_k)cov(X_i, X_j)}{var(X_i)var(X_j) - (cov(X_i, X_j))^2}
\end{split}
\end{equation}

As we want to nullify the effect of $X_k$ from both $X_i$ and $X_j$, the adjustment factor of $X_k$, $AF_k$ is only function of $X_i$ and $X_j$. The detailed steps are in Appendix \ref{app:feat_2}. By the similar manner, the combination of two features can be easily expanded to combination of $p+1$ $(p+1 \leq m)$ features. In this case equation \ref{eq:two_feat_1} becomes,

\begin{equation} 
	\label{eq:p_feat_1}
	\begin{split}
		cor(X_i, X_k +  AF_k) = 0 \\
		cor(X_{i+1}, X_k +  AF_k) = 0 \\
		cor(X_{i+2}, X_k +  AF_k) = 0  \\
		\makebox[11em]{\dotfill}\\
		\makebox[11em]{\dotfill}\\
		\makebox[11em]{\dotfill}\\
		cor(X_{i+p}, X_k +  AF_k) = 0 
	\end{split}
\end{equation}

Putting
\begin{equation}
	AF_k = a_iX_i+a_{i+1}X_{i+1}+a_{i+2}X_{i+2}+...+a_{i+p}X_{i+p}
	\label{eq:af_p}
\end{equation} 

in the equation \ref{eq:p_feat_1} and writing the $p$ equations in matrix form we get,

\begin{equation}
\scriptsize{\begin{split}
\begin{bmatrix} 
	var(X_i) & cov(X_i, X_{i+1}) & cov(X_i, X_{i+2}) & ... & cov(X_i, X_{i+p})\\ 
	cov(X_{i+1}, X_i) & var(X_{i+1}) & cov(X_{i+1}, X_{i+2}) & ... & cov(X_{i+1}, X_{i+p})\\  
	cov(X_{i+2}, X_i) & cov(X_{i+2}, X_{i+1}) & var(X_{i+2}) & ... & cov(X_{i+2}, X_{i+p})\\ 
	... & ... & ... & ... & ... \\
	... & ... & ... & ... & ... \\
	... & ... & ... & ... & ... \\
	cov(X_{i+j}, X_i) & cov(X_{i+j}, X_{i+1}) & cov(X_{i+j}, X_{i+2}) & ... & cov(X_{i+j}, X_{i+p})\\ 
	... & ... & ... & ... & ... \\
	... & ... & ... & ... & ... \\
	... & ... & ... & ... & ... \\
	cov(X_{i+p}, X_i) & cov(X_{i+p}, X_{i+1}) & cov(X_{i+p}, X_{i+2}) & ... & var(X_{i+p})\\ 
\end{bmatrix} 
\begin{bmatrix} 
	X_i \\
	X_{i+1} \\
	X_{i+2} \\
	... \\
	... \\
	... \\
	X_{i+j} \\
	... \\
	... \\
	... \\
	X_{i+p} \\
\end{bmatrix} 
=
\begin{bmatrix}
	cov(X_i, X_k) \\
	cov(X_{i+1}, X_k) \\
	cov(X_{i+2}, X_k) \\
	... \\
	... \\
	... \\
	cov(X_{i+j}, X_k) \\
	... \\
	... \\
	... \\
	cov(X_{i+p}, X_k) \\
\end{bmatrix}
\end{split}}
\end{equation}

By Cramer's rule,

\begin{equation}
	a_{i+j} = \frac{\det A_{j+1}}{\det A}
\end{equation}

where, $\scriptsize{A_{j+1} = \begin{bmatrix}
	var(X_i) & ...& cov(X_i, X_{i+j-1})  & cov(X_i, X_{k}) & ... & cov(X_i, X_{i+p})\\ 
cov(X_{i+1}, X_i) & ...& cov(X_{i+1}, X_{i+j-1}) & cov(X_{i+1}, X_{k}) & ... & cov(X_{i+1}, X_{i+p})\\  
cov(X_{i+2}, X_i)  & ...&  cov(X_{i+2}, X_{i+j-1}) &  cov(X_{i+2}, X_{k}) & ... & cov(X_{i+2}, X_{i+p})\\ 
... & ... &  ... & ... & ... & ... \\
... & ... &  ... & ... & ... & ... \\
... & ... &  ... & ... & ... & ... \\
cov(X_{i+j}, X_i)  & ... & cov(X_{i+j}, X_{i+j-1}) & cov(X_{i+j}, X_{k}) & ... & cov(X_{i+j}, X_{i+p})\\ 
... & ... &  ... & ... & ... & ... \\
... & ... &  ... & ... & ... & ... \\
... & ... &  ... & ... & ... & ... \\
cov(X_{i+p}, X_i)  & ... & cov(X_{i+p}, X_{i+j-1}) & cov(X_{i+p}, X_{k}) & ... & var(X_{i+p})\\ 
\end{bmatrix}}$

\vskip 1.5em
\hskip 2.3em and, $\scriptsize{A = \begin{bmatrix}
		var(X_i) & cov(X_i, X_{i+1}) & cov(X_i, X_{i+2}) & ... & cov(X_i, X_{i+p})\\ 
	cov(X_{i+1}, X_i) & var(X_{i+1}) & cov(X_{i+1}, X_{i+2}) & ... & cov(X_{i+1}, X_{i+p})\\  
	cov(X_{i+2}, X_i) & cov(X_{i+2}, X_{i+1}) & var(X_{i+2}) & ... & cov(X_{i+2}, X_{i+p})\\ 
	... & ... & ... & ... & ... \\
	... & ... & ... & ... & ... \\
	... & ... & ... & ... & ... \\
	cov(X_{i+j}, X_i) & cov(X_{i+j}, X_{i+1}) & cov(X_{i+j}, X_{i+2}) & ... & cov(X_{i+j}, X_{i+p})\\ 
	... & ... & ... & ... & ... \\
	... & ... & ... & ... & ... \\
	... & ... & ... & ... & ... \\
	cov(X_{i+p}, X_i) & cov(X_{i+p}, X_{i+1}) & cov(X_{i+p}, X_{i+2}) & ... & var(X_{i+p})\\ 
	\end{bmatrix}}$

\vskip 1em
Once we calculate all the $a_{i+j}$, where $j = 0, 1, 2, ..., p$, those values can be put back in equation \ref{eq:af_p} to get the adjustment factor for feature $X_j$ while calculating MCC shapley values.

\subsubsection{Algorithm of MCC Shapley Value Calculation}
For MCC shapley value calculation we used our adjustment factor in approximate shapley value calculation with Monte-Carlo sampling proposed by \cite{montecarlo} as original shapley value calculation is very time consuming and practically infeasible for large number of features. Algorithm \ref{alg:mccshapley} contains estimation of MCC shapley values. This algorithm is exactly same as the algorithm of approximate shapley value calculation written in the chapter 5 of the book by \cite{molnar2019} except line 9 and 10, where we add our novel adjustment factors to each of the features in coalitions excluding $x^{(i)}_{j}$ as $X_j$ is the feature of interest for calculation of MCC shapley values for $x^{(i)}$. It is to be noted that, as correlation is only restricted to numerical variables, the multi-collinearity correction is only applicable to the same. So, in step 9 and 10, $X_j$ and the features with which we add $AF$ must be numerical.

\begin{algorithm}
	\caption{Estimation of MCC Shapley Values}\label{alg:mccshapley}
	\begin{algorithmic}[1]
		\State \textbf{Output:} MCC Shapley value for the value of the j-th feature, $x^{(i)}_j$.
		\State \textbf{Required:} Number of iterations $M$, instance of interest $x^{(i)}$, feature index $j$, data matrix \textbf{X}, and machine learning model $f$
		\For {$m\gets 1, M$}
		\State Draw random instance $x^{(r)}$ from the data matrix X
		\State Choose a random permutation o of the feature values
		\State Order instance $x^{(i)}$: $x^{(i)}_{[o]} = (x^{(i)}_{(1)}, ..., x^{(i)}_{(j)}, ..., x^{(i)}_{(m)})$
		\State Order instance $x^{(r)}$: $x^{(r)}_{[o]} = (x^{(r)}_{(1)}, ..., x^{(r)}_{(j)}, ..., x^{(r)}_{(m)})$
		\State Construct two new instances adding adjustment factors to the feature values in coalitions i.e features belongs to instance $x^{(i)}$
		\State \hskip 2.1em $\bullet$ With feature j: $x_{+j} = (x^{(i)}_{(1)}+AF_{(1)}, ..., x^{(i)}_{(j-1)}+AF_{(j-1)},x^{(i)}_{(j)},x^{(r)}_{(j+1)}, ..., x^{(r)}_{(m)})$
		\State \hskip 2.1em $\bullet$ Without feature j: $x_{-j} = (x^{(i)}_{(1)}+AF_{(1)}, ..., x^{(i)}_{(j-1)}+AF_{(j-1)},x^{(r)}_{(j)},x^{(r)}_{(j+1)}, ..., x^{(r)}_{(m)})$
		\State Compute marginal contribution: $\phi_j^m = \hat{f}(x_{+j}) -\hat{f}(x_{-j}) $
		\EndFor
		\State Compute MCC Shapley value as the average: $\phi_j(x) = \frac{1}{M} \sum_{m=1}^{M}\phi_j^m$
	\end{algorithmic}
\end{algorithm}

\section{Results}
We experimented on those datasets where at least two features have moderate to high correlation values between them for the MCC shapley value calculation for indivudual features and combination of two features. The result is very intuitive to understand with the presence of strong multi-collinearity (correlation value $\approx$ 1). This statement will become clearer once we show and explain the results in this section. We mainly focused to calculate MCC shapley values with individual and combination of two features. The other combinations can be easily calculated based on the matrix form shown in the section \ref{sec:mcc2}.

\subsection{Dataset - House Prices}
This dataset from \cite{houseprice} presents a regression problem where given the attributes of a house, the prediction of the price of the house to be predicted. We did the pre-processing which includes handling missing values and creation of dummy variables for the categorical variables to make the dataset prepared for model fitting with the final 331 predictors.

\subsubsection{Results of MCC Shapley Values for Individual Features}
\label{sec:res_indv}

As the dataset with features with strong multi-collinearity (correlation $\approx$ 1) is hard to find, we break our results into two scenarios, in first scenario we add one (or two) artificially created variables with correlation $\approx$ 1 with any one (or two) of the features to understand the effect of multi-collinearity correction. Once we understand this, in the second scenario we shall observe the effect of the same with two real features with high correlation. We refrain ourselves from fitting linear models as the estimate of the coefficients become unstable due to the presence of multi-collinearity. From now on we term shapley values with and without multi-collinearity as MCC-SV and NMCC-SV respectively.

In scenario 1, we picked one numerical feature \texttt{MiscVal} as our feature of interest because \texttt{MiscVal} has very low correlation ($\approx$ 0) with the other numerical features. We shall explain the reason behind doing so after we show the result. We created one highly correlated variable (correlation $\approx$ 1) with \texttt{MiscVal} artificially and name it \texttt{MiscVal\_corr}. This makes \texttt{MiscVal} and \texttt{MiscVal\_corr} the equally important feature to the output feature  \texttt{SalePrice}. Table \ref{tab:hp_indv_s1} shows the shapley values with and without multi-collinearity correction for a randomly picked data point for \texttt{MiscVal} feature. In the table \textbf{Without Presence of Artifically Created Variable} and \textbf{With Presence of Artifically Created Variable} mean the situations where models are trained with the original 331 predictors and 331+1 artificially created feature (\texttt{MiscVal\_corr}) respectively. From Table \ref{tab:hp_indv_s1} it is seen that with the presence of \texttt{MiscVal\_corr} the NMCC shapley values have sliced to half of the NMCC shapley values without the presence of \texttt{MiscVal\_corr} for all the models. This is very intuitive because as both \texttt{MiscVal} and \texttt{MiscVal\_corr} are the equally important features to the output feature and this is the reason we created the artificial feature with correlation $\approx$ 1. But this reduction of the shapley values with the presence of correlated variable is highly unwanted as stated earlier, because if \texttt{MiscVal} is an important variable to the output then the importance of it reduces down due to the presence of a similar important variable. It is also seen from the table that due to our novel multi-collinearity correction the MCC shapley values of both \texttt{MiscVal} and \texttt{MiscVal\_corr} increased back to the earlier NMCC shapley values without the presence of \texttt{MiscVal\_corr}. Under \textbf{Without Presence of Artifically Created Variable} column header along with \textbf{NMCC-SV of MiscVal} there is another column \textbf{MCC-SV of MiscVal} which calculates the MCC shapley value of \texttt{Miscval} without presence of \texttt{MiscVal\_corr}. We see that the values of both the columns are almost same, because the correlation of the other features with \texttt{MiscVal} is almost 0. This is the reason we chose this type of feature to have a sanity check of the performance of our multi-collinearity correction.

\begin{table}
	\centering
	\arrayrulecolor{black}
	\begin{adjustbox}{width=\columnwidth,center}
	\begin{tabular}{|l|p{2.2cm}|p{2.0cm}|p{2.2cm}|p{2.7cm}|p{2.2cm}|p{2.7cm}|}
		\hline
		\multirow{2}{*}{\textbf{Model}}    & \multicolumn{2}{p{3.9cm}|}{\textbf{Without Presence of Artifically Created Variable}} & \multicolumn{4}{c!{\color{black}\vrule}}{\textbf{With Presence of Artifically Created Variable}}       \\ 
		\cline{2-7}
		& \textbf{NMCC-SV of MiscVal}    &    \textbf{MCC-SV of MiscVal}                          & \textbf{NMCC-SV of MiscVal} & \textbf{NMCC-SV of MiscVal\_corr} & \textbf{MCC-SV of MiscVal} & \textbf{MCC-SV of MiscVal\_corr}  \\ 
		\hline
		Decision Tree             & -261.4 $\pm$ 5.3   & -260.2 $\pm$ 5.3                                             & -129.3 $\pm$ 5.1                  & -128.3 $\pm$ 4.8                       & -263.1 $\pm$ 5.4                   & -262.9 $\pm$ 4.7                         \\ 
		\hline
		Random Forest             & -210.9 $\pm$ 1.4       & -209.8 $\pm$ 1.4                                       & -107.8 $\pm$ 1.1                  & -102.4 $\pm$ 1.2                       & -213.5 $\pm$ 0.9                   & -210.2 $\pm$ 1.3                         \\ 
		\hline
		Gradient Boosting        & -273.2 $\pm$ 1.1     & -272.1 $\pm$ 1.1                                         & -138.1 $\pm$ 1.4                  & -137.3 $\pm$ 1.2                       & -276.3 $\pm$ 1.8                   & -275.2 $\pm$ 1.0  \\
		\hline
		Extreme Gradient Boosting  & -265.9 $\pm$ 1.2    & -263.8 $\pm$ 1.2                                          & -135.2 $\pm$ 1.2                  & -131.7 $\pm$ 1.3                       & -265.0 $\pm$ 0.9                   & -264.2 $\pm$ 1.1  \\ 
		\hline
		Support Vector Regression  & -112.6 $\pm$ 0.4        & -113.9 $\pm$ 0.4                                       & -53.4 $\pm$ 0.3                  & -55.6 $\pm$ 0.1                       & -111.0 $\pm$ 0.1                   & -115.7 $\pm$ 0.2  \\
		\hline
	\end{tabular}
	\end{adjustbox}
	\arrayrulecolor{black}
	\caption{Shapley Values with and without Multi-collinearity Correction for a randomly picked data point for \texttt{MiscVal} Feature. These values are created for a Monte-Carlo simulation with 10000 iterations. }
	\label{tab:hp_indv_s1}
\end{table}

In Scenario 2, We picked feature \texttt{1stFlrSF} and we see that \texttt{1stFlrSF} has high correlation (0.82) with \texttt{TotalBsmtSF}. To understand the effect of the multi-collinearity correction, in one setting we remove highly correlated features with \texttt{1stFlrSF} and build the model to calculate the shapley values. Then in the second setting we use all the 331 features of the dataset to build the model and compare the MCC and NMCC shapley values. Table \ref{tab:hp_indv_s2} shows shapley values with and without Multi-collinearity Correction for a randomly picked data point for \texttt{1stFlrSF} Feature. In the first setting \textbf{Without Presence of TotalBsmtSF} we have compared between MCC and NMCC shapley values. As we build the model removing the highly correlated features with \texttt{1stFlrSF} we see that there is not much difference between NMCC and MCC shapley values, but there is one interesting pattern which is that the MCC shapley values are always slightly greater than the NMCC shapley values. Though we removed the highly correlated features while building the model there were some features (e.g. \texttt{BsmtFinSF1, GarageArea, BsmtUnfSF, WoodDeckSF,} etc.) having moderate/low correlation with \texttt{1stFlrSF}, thus they reduced the NMCC shapley values to a little extent. But with the presence of the highly correlated feature \texttt{TotalBsmtSF}, we see that the NMCC shapley values has reduced down a lot, but the reduction is not roughly $50\%$ like scenario 1, because here the correlation between \texttt{TotalBsmtSF} and \texttt{1stFlrSF} is not 1. With the use of our multi-collinearity correction the MCC shapley values increased and roughly matches with the MCC shapley values calculated from the model built on \textbf{Without Presence of TotalBsmtSF} in dataset. The match is not very close because we are comparing the the shapley values of two different models and in one of the model \texttt{TotalBsmtSF} feature is not available.

\begin{table}
	\centering
	\arrayrulecolor{black}
	\begin{adjustbox}{width=\columnwidth,center}
		\begin{tabular}{|l|p{2.2cm}|p{2.2cm}|p{2.2cm}|p{2.2cm}|}
			\hline
			\multirow{2}{*}{\textbf{Model}}    & \multicolumn{2}{p{4.4cm}|}{\textbf{Without Presence of TotalBsmtSF}}& \multicolumn{2}{p{4.4cm}|}{\textbf{With Presence of all of 331 features}}       \\ 
			\cline{2-5}
			& \textbf{NMCC-SV of 1stFlrSF}         &  \textbf{MCC-SV of 1stFlrSF}                  & \textbf{NMCC-SV of 1stFlrSF}  & \textbf{MCC-SV of 1stFlrSF}   \\ 
			\hline
			Decision Tree             & 2721.1 $\pm$ 5.1                & 2832.2 $\pm$ 5.0                              & 1932.8 $\pm$ 5.6                                  & 2841.7 $\pm$ 5.2                                        \\ 
			\hline
			Random Forest             & 2502.3 $\pm$ 1.1              & 2715.4 $\pm$ 1.6                               & 1781.3 $\pm$ 1.2                                       & 2700.4 $\pm$ 1.5                                          \\ 
			\hline
			Gradient Boosting        & 2657.7 $\pm$ 0.9                    & 2919.1 $\pm$ 1.1                       & 1699.1 $\pm$ 0.9                                        & 2933.4 $\pm$ 1.0                     \\
			\hline
			Extreme Gradient Boosting  & 3356.2 $\pm$ 1.4              & 3612.6 $\pm$ 0.9                               & 2134.5 $\pm$ 1.5                                     & 3597.8 $\pm$ 1.1                     \\ 
			\hline
			Support Vector Regression  & 1745.8 $\pm$ 0.5            & 1983.7 $\pm$ 0.4                                & 1244.9 $\pm$ 0.7                                        & 1998.4 $\pm$ 0.8                     \\
			\hline
		\end{tabular}
	\end{adjustbox}
	\arrayrulecolor{black}
	\caption{Shapley Values with and without Multi-collinearity Correction for a randomly picked data point for \texttt{1stFlrSF} Feature. These values are created for a Monte-Carlo simulation with 10000 iterations. }
	\label{tab:hp_indv_s2}
\end{table}

\subsubsection{Results of MCC Shapley Values for Combination of two Features}
For combination of two features, for scenario 1, We picked two features \texttt{MiscVal} and \texttt{3SsnPorch} who are almost uncorrelated with the other features. We created two highly correlated artificial features and name those \texttt{MiscVal\_corr} and \texttt{3SsnPorch\_corr}. Table \ref{tab:hp_comb2_s1} shows the result of the effect of Multi-collinearity correction for the combination of two features which is analogous of Table \ref{tab:hp_indv_s1} for individual features. From the result it is seen that the NMCC shapley values sliced to half due to presence of the perfectly correlated features, but the multi-collinearity correction factor helped those features to get back the actual values i.e. MCC shapley values. Also, as \texttt{MiscVal} and \texttt{3SsnPorch} are almost uncorrelated with other features, NMCC and MCC shapley values are almost same without the presence of artificially created variable across all the models.

\begin{table}
	\centering
	\arrayrulecolor{black}
	\begin{adjustbox}{width=\columnwidth,center}
		\begin{tabular}{|l|p{3.1cm}|p{3.0cm}|p{3.1cm}|p{3.0cm}|}
			\hline
			\multirow{2}{*}{\textbf{Model}}    & \multicolumn{2}{p{6.1cm}|}{\textbf{Without Presence of Artifically Created Variable}} & \multicolumn{2}{p{6.1cm}|}{\textbf{With Presence of Artifically Created Variable}}       \\ 
			\cline{2-5}
			& \textbf{NMCC-SV of the combination of MiscVal and 3SsnPorch}    &    \textbf{MCC-SV of the combination of MiscVal and 3SsnPorch}                          & \textbf{NMCC-SV of the combination of MiscVal and 3SsnPorch} & \textbf{MCC-SV of the combination of MiscVal and 3SsnPorch}   \\ 
			\hline
			Decision Tree             & 417.2 $\pm$ 4.9   & 415.6 $\pm$ 5.0                                             & 211.4 $\pm$ 5.1                                & 419.6 $\pm$ 4.8                                         \\ 
			\hline
			Random Forest             & 323.6 $\pm$ 1.9       & 327.8 $\pm$ 1.8                                       & 163.8 $\pm$ 1.9                  & 330.9 $\pm$ 2.0                                          \\ 
			\hline
			Gradient Boosting        & 374.3 $\pm$ 1.4     & 376.1 $\pm$ 1.7                                         & 185.0 $\pm$ 1.3                  & 370.7 $\pm$ 1.5                        \\
			\hline
			Extreme Gradient Boosting  & 289.5 $\pm$ 1.1    & 292.5 $\pm$ 1.4                                          & 149.7 $\pm$ 1.7                  & 289.1 $\pm$ 1.6                      \\ 
			\hline
			Support Vector Regression  & 134.5 $\pm$ 0.7        & 145.3 $\pm$ 0.8                                       & 67.3 $\pm$ 0.9                  & 149.2 $\pm$ 0.6                   \\
			\hline
		\end{tabular}
	\end{adjustbox}
	\arrayrulecolor{black}
	\caption{Shapley Values with and without Multi-collinearity Correction for a randomly picked data point for the combination of \texttt{MiscVal} and \texttt{3SsnPorch} features. These values are created for a Monte-Carlo simulation with 10000 iterations. }
	\label{tab:hp_comb2_s1}
\end{table}

For scenario 2, we just compared the shapley values with and without multi-collinearity correction for a combination of two features i.e. \texttt{1stFlrSF} and \texttt{2ndFlrSF}. Table \ref{tab:hp_comb2_s2} shows the result from which it is seen that as \texttt{1stFlrSF} and \texttt{2ndFlrSF} have moderate to strong correlation with other features, due to correction the MCC shapley values increase with respect to their NMCC shapley values counter-parts. 

\begin{table}[ht]
	\centering
	\arrayrulecolor{black}
		\begin{tabular}{|l|p{4cm}|p{4cm}|}
			\hline
			\textbf{Model} & \textbf{NMCC-SV of combination of 1stFlrSF and 2ndFlrSF} & \textbf{MCC-SV of combination of 1stFlrSF and 2ndFlrSF} \\
			\hline
			Decision Tree  & 3321.5 $\pm$ 3.2 & 4610.3 $\pm$ 3.3 \\
			\hline
			Random Forest  & 2895.4 $\pm$ 1.4 & 3767.6 $\pm$ 1.6 \\
			\hline
			Gradient Boosting  & 3006.7 $\pm$ 1.3 & 4209.2 $\pm$ 1.3 \\
			\hline
			Extreme Gradient Boosting  & 3209.1 $\pm$ 0.9 & 4479.9 $\pm$ 1.0 \\
			\hline
			Support Vector Regression  & 3877.0 $\pm$ 0.7 & 5003.6 $\pm$ 0.9 \\
			\hline
			
		\end{tabular}
	\arrayrulecolor{black}
	\caption{Shapley Values with and without Multi-collinearity Correction for a randomly picked data point for the combination of \texttt{1stFlrSF} and \texttt{2ndFlrSF} features. These values are created for a Monte-Carlo simulation with 10000 iterations. }
	\label{tab:hp_comb2_s2}
\end{table}

\subsection{Dataset - Default of Credit Card Clients}
This data set by \cite{Yeh2009TheCO} aimed at the case of customers' default payments in Taiwan and predicts probability of default of clients. For this data set, to compare results between with and without multi-collinearity correction for individual or combination of features, we mainly focused on the real features like scenario 2 in the earlier section. As this a classification problem the shapley values are produced in logit terms. Table \ref{tab:dp_indv_s2} and Table \ref{tab:dp_comb_s2} shows the effect of multi-collinearity correction for individual and combination of two features respectively. As there are correlation with the features it is seen that for almost all the cases across all the models the shapely values increase after multi-collinearity correction.

\begin{table}[H]
	\centering
	\arrayrulecolor{black}
	\begin{adjustbox}{width=\columnwidth,center}
		\begin{tabular}{|l|p{1.5cm}|p{2.5cm}|p{2.4cm}|p{1.5cm}|p{2.5cm}|p{2.4cm}|}
			\hline
			\multirow{2}{*}{\textbf{Model}}    & \multicolumn{3}{p{6.1cm}|}{\textbf{NMCC Shapley Values}} & \multicolumn{3}{p{6.1cm}|}{\textbf{MCC Shapley Value}}       \\ 
			\cline{2-7}
			& \textbf{PAY\_6}    &    \textbf{BILL\_AMT6}    & \textbf{PAY\_AMT6} &\textbf{PAY\_6}    &    \textbf{BILL\_AMT6}    & \textbf{PAY\_AMT6}   \\ 
			\hline
			Decision Tree             & 3.7 $\pm$ 0.4   & 4.9 $\pm$ 0.3                                             & 1.7 $\pm$ 0.2                                & 6.6 $\pm$ 0.3      & 0.5 $\pm$ 0.2                                & 2.6 $\pm$ 0.3                                    \\ 
			\hline
			Random Forest             & 3.2 $\pm$ 0.1       & 4.0 $\pm$ 0.1                                       & 1.8 $\pm$ 0.1                  & 6.0 $\pm$ 0.1                   & 7.4 $\pm$ 0.1                                & 3.0 $\pm$ 0.1                        \\ 
			\hline
			Gradient Boosting        & 2.9 $\pm$ 0.1     & 5.1 $\pm$ 0.1                                         & 2.0 $\pm$ 0.1                  & 5.9 $\pm$ 0.1      & 9.8 $\pm$ 0.1                                & 3.6 $\pm$ 0.1                   \\
			\hline
			Extreme Gradient Boosting  & 4.5 $\pm$ 0.1    & 4.3 $\pm$ 0.1                                          & 2.1 $\pm$ 0.1                  & 7.1 $\pm$ 0.1   & 8.1 $\pm$ 0.1                                & 3.6 $\pm$ 0.1                    \\ 
			\hline
		\end{tabular}
	\end{adjustbox}
	\arrayrulecolor{black}
	\caption{Shapley Values with and without Multi-collinearity Correction for a randomly picked data point for three individual features i.e. \texttt{PAY\_6, BILL\_AMT6} and \texttt{PAY\_AMT6}. These values are created for a Monte-Carlo simulation with 10000 iterations. }
	\label{tab:dp_indv_s2}
\end{table}

\begin{table}
	\centering
	\arrayrulecolor{black}
	\begin{adjustbox}{width=\columnwidth,center}
		\begin{tabular}{|l|p{1.8cm}|p{2.6cm}|p{1.8cm}|p{2.6cm}|}
			\hline
			\multirow{2}{*}{\textbf{Model}}    & \multicolumn{2}{p{4.4cm}|}{\textbf{NMCC Shapley Values}} & \multicolumn{2}{p{4.4cm}|}{\textbf{MCC Shapley Value}}       \\ 
			\cline{2-5}
			& \textbf{PAY\_2 \& PAY\_3}    &    \textbf{BILL\_AMT5 \& BILL\_AMT6}    & \textbf{PAY\_2 \& PAY\_3}    &    \textbf{BILL\_AMT5 \& BILL\_AMT6}    \\ 
			\hline
			Decision Tree             & 6.1 $\pm$ 0.2   & 2.9 $\pm$ 0.3                                             & 10.7 $\pm$ 0.3                                & 5.5 $\pm$ 0.2                                \\ 
			\hline
			Random Forest             & 5.1 $\pm$ 0.1       & 3.0 $\pm$ 0.1                                       & 9.1 $\pm$ 0.1                  & 5.6 $\pm$ 0.1                                   \\ 
			\hline
			Gradient Boosting        & 4.8 $\pm$ 0.1     & 3.4 $\pm$ 0.1                                         & 9.0 $\pm$ 0.1                  & 6.0 $\pm$ 0.1                 \\
			\hline
			Extreme Gradient Boosting  & 5.3 $\pm$ 0.1    & 3.9 $\pm$ 0.1                                          & 9.9 $\pm$ 0.1                  & 7.3 $\pm$ 0.1                 \\ 
			\hline
		\end{tabular}
	\end{adjustbox}
	\arrayrulecolor{black}
	\caption{Shapley Values with and without Multi-collinearity Correction for a randomly picked data point for two sets of combination of two features i.e. \texttt{PAY\_2 \& PAY\_3} and \texttt{BILL\_AMT5 \& BILL\_AMT6}. These values are created for a Monte-Carlo simulation with 10000 iterations. }
	\label{tab:dp_comb_s2}
\end{table}

We performed one additional experiment where we compared the execution time between the MCC and NMCC shapley value calculation. This experiment is done in a machine with 2.6 GHz Intel Core i7 and 8 GB available RAM. Table \ref{tab:time} shows the comparison in execution time of the Shapley Values with and without multi-collinearity correction. This result is produced with a \texttt{Random Forest} model which is trained on default parameters and the number of iterations of Monte-Carlo simulation is 10,000. From the table it is seen that introduction of the correlation adjustment factor does not almost have any effect in the execution time for calculating shapley values.

\begin{table}
	\centering
	\arrayrulecolor{black}
	\begin{tabular}{lcc}
		\hline
		\textbf{Feature Size} & \textbf{NMCC-SV}(sec) & \textbf{MCC-SV}(sec) \\
		\hline
		$\approx$ 10  & 0.3 $\pm$ 0.01 & 0.3 $\pm$ 0.03 \\
	
		$\approx$ 100  & 0.9 $\pm$ 0.02 & 1.0 $\pm$ 0.03 \\
		
		$\approx$ 1000  & 4.7 $\pm$ 0.01 & 4.9 $\pm$ 0.01 \\
		
		$\approx$ 10000  & 17.3 $\pm$ 0.01 & 18.1 $\pm$ 0.01 \\
		\hline
		
	\end{tabular}
	\arrayrulecolor{black}
	\caption{Comparison of Shapley Value Calculation with and without correlation adjustment. This result is produced with \texttt{Random Forest} model and 10000 Monte-Carlo iterations.}
	\label{tab:time}
\end{table}

\section{Conclusion}
This paper shows that the novel multi-collinearity correction factor with shapley values helps to interpret features more accurately with the presence of moderate to high correlation within features in a data. Our algorithm is tested on multiple datasets and multiple models to prove it's efficacy. For better intuitive understanding we analysed the effect of our novel multi-collinearity correction factor with the presence of both artificially created features and real features and concluded it's effectiveness from different perspective. Finally we analysed the effect of the multi-collinearity correction factor in execution time and concluded that with the presence of multi-collinearity correction factor, the execution time is almost same compared to the calculation of shapley values without multi-collinearity correction factor.

\bibliography{references}

\clearpage
\appendix
\section{Detailed Steps for the Calculation of Adjustment Factor for Individual Features}
\label{app:feat_1}

\begin{equation*}
\begin{split}
cor(X_j, X_k +  AF_k) = 0 \\
\implies cov(X_j, X_k +  AF_k) = 0
\end{split}
\label{eq:indv_1}
\end{equation*}

Putting $AF_k = aX_j$ in the above equation we get,

\begin{align*}
cov(X_j,X_k+aX_j) = 0 \\
\implies cov(X_j, X_k) + a \times cov(X_j, X_j) = 0 \\
\implies a = - \frac{cov(X_j, X_k)}{var(X_j)}
\end{align*}

Putting the value of $a$, we get, $$AF_k =  - \frac{cov(X_j, X_k)}{var(X_j)}X_j$$

\section{Detailed Steps for the Calculation of Adjustment Factor for Combination of Two Features}
\label{app:feat_2}

Putting $AF_k = aX_i+bX_j$ in the equation \ref{eq:two_feat_1} we get,

\begin{equation}
	\begin{split}
		cor(X_i,X_k+aX_i+bX_j) = 0 \\
		\implies cov(X_i,X_k+aX_i+bX_j) = 0 \\
		\implies cov(X_i, X_k) + a \times cov(X_i, X_i) + b \times cov(X_i, X_j) = 0 \\
		\implies cov(X_i, X_k) + a \times var(X_i)+ b \times cov(X_i, X_j) = 0 
	\end{split}
\end{equation}

and,

\begin{equation}
	\begin{split}
		cor(X_j,X_k+aX_i+bX_j) = 0 \\
		\implies cov(X_j,X_k+aX_i+bX_j) = 0 \\
		\implies cov(X_j, X_k) + a \times cov(X_j, X_i) + b \times cov(X_j, X_j) = 0 \\
		\implies cov(X_j, X_k) + a \times cov(X_i, X_j)+ b \times var(X_j) = 0 
	\end{split}
\end{equation}

Solving above two equations for $a$ and $b$, we get the values shown in equation \ref{eq:two_feat_2}.

\end{document}